\documentclass[letterpaper, 10 pt, conference]{ieeeconf}  
\makeatletter
\def\endfigure{\end@float}
\def\endtable{\end@float} 
\makeatother

\IEEEoverridecommandlockouts                              

\usepackage{cite}
\usepackage{graphics} 
\usepackage{epsfig} 
\usepackage{times} 
\usepackage{amsmath} 
\usepackage{amssymb}  
\usepackage[utf8]{inputenc}
\usepackage{multirow} 
\usepackage{graphicx}
\usepackage{array}
\usepackage{multirow}
\usepackage{diagbox}
\usepackage{hhline}
\usepackage{arydshln}
\usepackage{booktabs}
\usepackage{xcolor}
\usepackage{url}
\usepackage{hyperref}

\renewcommand{\baselinestretch}{0.99}
\usepackage[caption=false]{subfig}

\title{\LARGE \bf
AF-RLIO: Adaptive Fusion of Radar-LiDAR-Inertial Information for Robust Odometry in Challenging Environments
}

\author{Chenglong Qian$^{1,\dagger}$, Yang Xu$^{2,\dagger}$, Xiufang Shi$^{1}$, Jiming Chen$^{2}$, and Liang Li$^{2,*}$
\thanks{$^{1}$College of Information Engineering, Zhejiang University of Technology, Hangzhou 310023, China.
$^{2}$College of Control Science and Engineering, Zhejiang University, Hangzhou 310027, China. \texttt{liang.li@zju.edu.cn}}
\thanks{This work was supported by National Natural Science Foundation of China (NSFC)  under Grants 62203383, U23A20326, 62372414, and 62403420, the National Postdoctoral Program for Innovative Talents under Grant BX20240317, and the China Postdoctoral Science Foundation under Grant 2024M752792. ($*:$ Corresponding author, $\dagger$: equal contribution.)}
}

\begin{document}

\maketitle
\thispagestyle{empty}
\pagestyle{empty}

\begin{abstract}

In robotic navigation, maintaining precise pose estimation and navigation in complex and dynamic environments is crucial. However, environmental challenges such as smoke, tunnels, and adverse weather can significantly degrade the performance of single-sensor systems like LiDAR or GPS, compromising the overall stability and safety of autonomous robots. To address these challenges, we propose AF-RLIO: an adaptive fusion approach that integrates 4D millimeter-wave radar, LiDAR, inertial measurement unit (IMU), and GPS to leverage the complementary strengths of these sensors for robust odometry estimation in complex environments. Our method consists of three key modules. Firstly, the pre-processing module utilizes radar data to assist LiDAR in removing dynamic points and determining when environmental conditions are degraded for LiDAR. Secondly, the dynamic-aware multimodal odometry selects appropriate point cloud data for scan-to-map matching and tightly couples it with the IMU using the Iterative Error State Kalman Filter. Lastly, the factor graph optimization module balances weights between odometry and GPS data, constructing a pose graph for optimization. The proposed approach has been evaluated on datasets and tested in real-world robotic environments, demonstrating its effectiveness and advantages over existing methods in challenging conditions such as smoke and tunnels. Furthermore, we open source our code at \href{https://github.com/QCL0920/AF-RLIO.git}{https://github.com/NeSC-IV/AF-RLIO.git} to benefit the research community.

\end{abstract}

\section{Introduction}
Robust and reliable odometry is crucial for autonomous robots when entering challenging environments. In complex environments such as smoke-filled areas or narrow tunnels, LiDAR-based fusion algorithms face significant challenges in simultaneous localization and mapping (SLAM) \cite{ebadi2022present,lee2024lidar,bijelic2018benchmark}.
Particles like smoke and dust degrade the quality of LiDAR scans, severely impairing point cloud registration \cite{bijelic2018benchmark}. The increased noise and uncertainty in sensor data in these environments render odometry systems that rely solely on LiDAR and inertial measurement units (IMUs) less accurate. Additionally, in narrow or enclosed environments, such as tunnels, dense urban areas, and canyons, maintaining reliable Global Positioning System (GPS) signals becomes challenging, and GPS outliers can lead to system failures\cite{liu2021optimization, nubert2022graph,wang2024end}. Therefore, to achieve accurate odometry estimation under extreme weather and challenging conditions, selecting more robust sensors and implementing appropriate multi-modal fusion is critical.

Compared to LiDAR, 4D millimeter-wave radar has a longer wavelength and is less affected by adverse weather conditions such as rain, snow, and fog \cite{harlow2023new}.

Additionally, its ability to sense over greater distances and capture Doppler information can enhance the robustness of odometry estimates in environments with similar geometric features. However, millimeter-wave radar point clouds are sparser and less stable than those generated by LiDAR. 

\begin{figure}
    \centering
    \subfloat[\centering Platform]{
        \includegraphics[width=0.9 \linewidth]{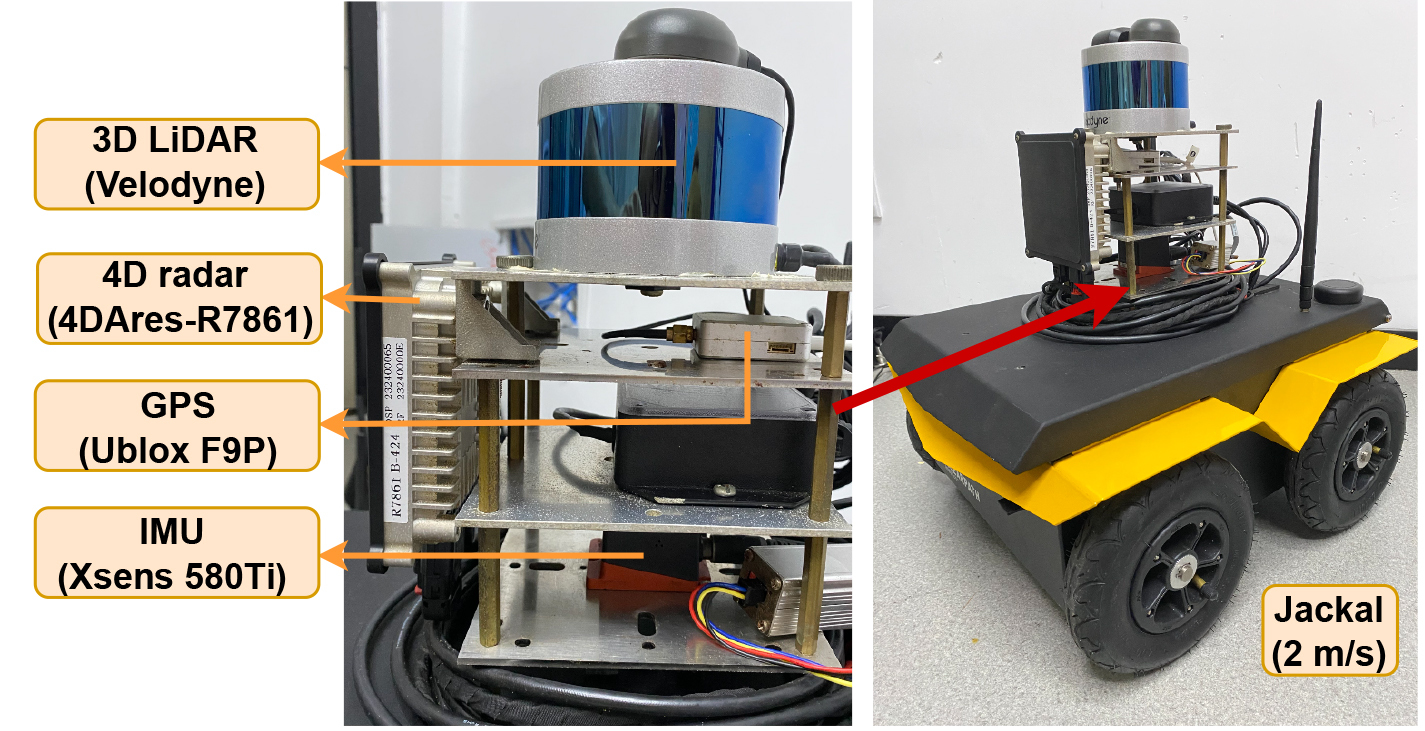}
        \label{Platform}
    }
    \quad
    \subfloat[\centering Odometry and mapping in smoke environment]{
        \includegraphics[width=0.9 \linewidth]{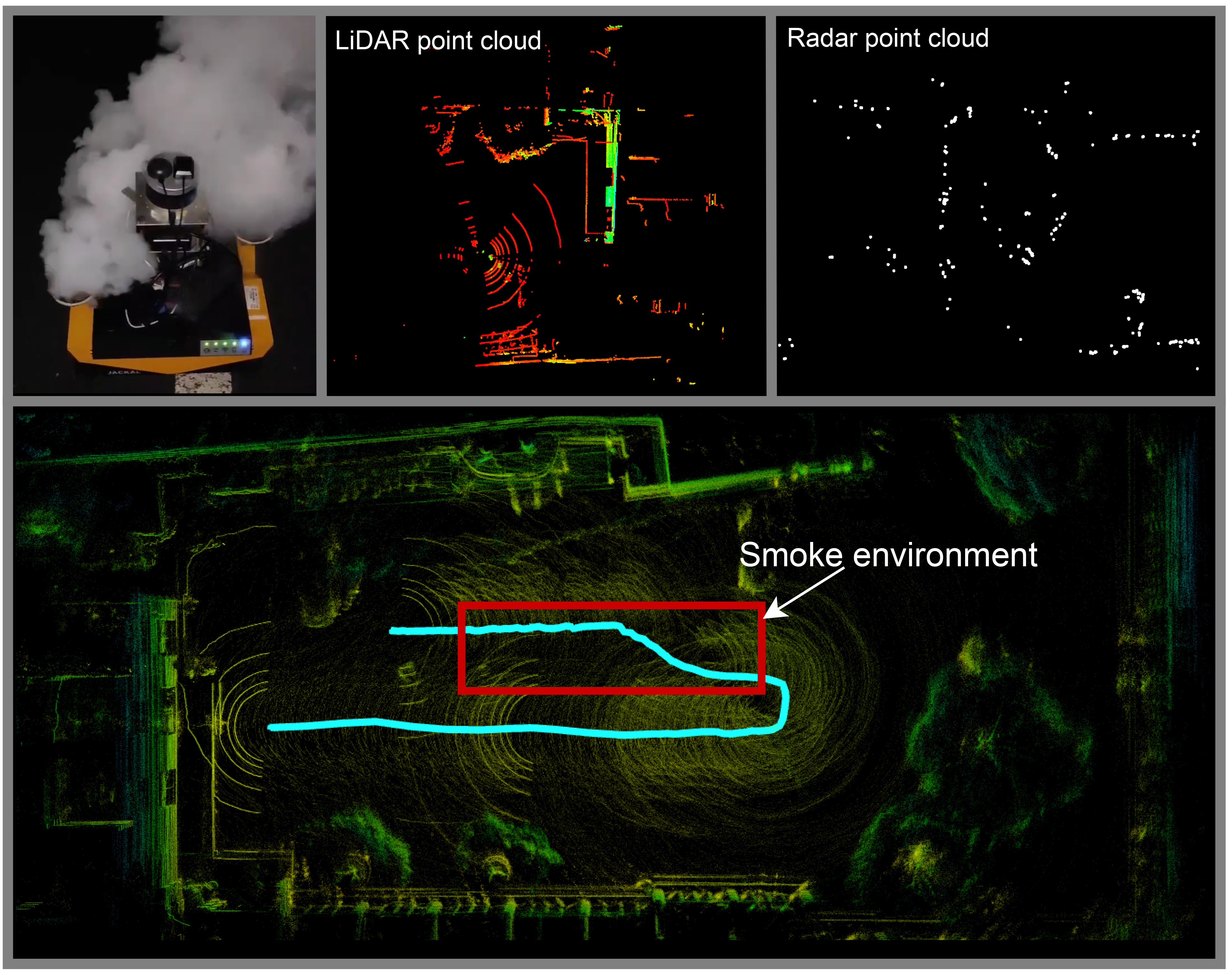}
        \label{Odometry and mapping in smoke environment}
    }
    \caption{(a) The test platform of the proposed system: a Jackal car with a speed of about 2m/s. (b) Display of LiDAR and radar point clouds in the smoke environment and verification of AF-RLIO adaptive fusion of radar data to complete odometry and mapping.}
    \label{Real_Experiment}
    \vspace{-5mm}
\end{figure}

To this end, we propose an IMU-centric multi-sensor fusion approach. Our multimodal fusion strategy employs radar point clouds to assist LiDAR in removing dynamic obstacles when LiDAR point clouds are effective, while tightly fusing LiDAR and IMU data to ensure reliable odometry. In environments where LiDAR performance degrades, the system adapts by switching to a tight fusion of millimeter-wave radar and IMU information to prevent odometry divergence.

To summarize, our key contributions are:
\begin{itemize}

    \item 
    We propose an adaptive multi-modal odometry system that can selectively couple radar and LiDAR with IMU tightly within the IESKF framework, followed by a loose fusion of weighted GPS information using pose graph optimization. This system fully leverages the strengths of each sensor under varying environmental conditions, enabling robust and accurate localization, even in challenging conditions where LiDAR or GPS is degraded.
    \item We design a robust GPS outlier detection method that utilizes radar-estimated ego velocity and dynamic-aware multimodal odometry to identify GPS outliers. The system smoothly adjusts GPS weights to mitigate the impact of outliers, improving overall system reliability.
    \item Extensive experiments were conducted on multiple datasets and in real-world scenarios, demonstrating the accuracy and robustness of the proposed algorithm. 

\end{itemize}

\section{Related Work}

\subsection{LiDAR-Inertial Odometry}

The LiDAR-inertial odometry (LIO) can be primarily categorized into filter-based\cite{qin2020lins,xu2021fast,xu2022fast,wu2024lio,bai2022faster} and optimization-based methods\cite{ye2019tightly,shan2020lio,li2021towards,zhao2021super}. 

Filter-based methods, such as the Extended Kalman Filter, typically update the system state vector and covariance to fuse new LiDAR and IMU measurements in real-time recursively and smoothly.
For instance, FAST-LIO2 \cite{xu2022fast} employs the Iterative Error State Kalman Filter (IESKF) without the need to refine map points or robot states through an optimizer. It registers points directly to the map without feature extraction, fully leveraging environmental information and adapting to emerging LiDAR technologies. 
On the other hand, optimization-based methods formulate the LIO problem as a nonlinear optimization, aiming to minimize the residuals between predicted and observed measurements over a sliding window or the entire trajectory. 

However, LiDAR is susceptible to environmental factors, and in smoke or extreme weather conditions, it may fail to produce registered frames \cite{ebadi2023present}. Even constrained by IMU, it may fail to maintain long-term stable and reliable odometry.

\subsection{Radar-Inertial Odometry}
In recent years, radar-based SLAM methods have gained increasing attention in robotics, particularly in autonomous driving \cite{li20234d,lim2023orora,burnett2021we,huai2024snail,choi2023msc,cen2018precise,cen2019radar}. 
These methods leverage the unique advantages of radar sensors, such as their robustness to adverse weather conditions (\textit{e.g.}, fog, rain, and dust) that typically degrade the performance of LiDAR and camera-based systems \cite{lu2020see}\cite{guan2020through}.
It also accelerates the development of radar inertial odometry (RIO), which involves the research of radar scan matching \cite{kubelka2024we} and multi-sensor fusion.
In EKF-based RIO methods \cite{michalczyk2022tightly,doer2020ekf,michalczyk2023multi}, inertial data, radar-based ego-velocity estimates, and radar point cloud registration are tightly integrated to estimate the 3D position and orientation of the robots. Additionally, inspired by optimization-based LIO methods, researchers formulate the RIO problem as a factor graph, enabling joint optimization of pose estimates over a sliding window or the entire trajectory \cite{kramer2020radar} \cite{kramer2021radar}.

\subsection{Radar-LiDAR-Inertial Odometry}

To fully leverage the strengths of both LiDAR and radar, recent research has focused on fusing these two sensors for localization and mapping\cite{fritsche2018fusing,yin2021rall,mielle2019comparative}. In scenarios where LiDAR degrades due to fog, dust, or low light, radar can continue to provide reliable measurements. Conversely, LiDAR's high-resolution data can refine the map and improve localization in clear conditions. For example, DR-LRIO \cite{nissov2024degradation} employs graph optimization to tightly couple LiDAR, radar, and IMU. It integrates radar-estimated linear velocity and poses difference estimates into the factor graph, enabling robust odometry in degraded environments. However, the long-term fusion of radar data may reduce overall localization accuracy due to the relatively large radar error. The approach in \cite{noh2024adaptivelidarradarfusionoutdoor} selects points for odometry estimation by identifying LiDAR-aware degraded points in the LiDAR point cloud. However, solely relying on radar or LiDAR, without proprioceptive sensors, like IMU and GPS, still poses significant challenges in achieving robust localization.

Therefore, this paper proposes a selective fusion approach that adaptively utilizes different exteroceptive sensors based on environmental conditions to handle adverse environmental changes, ensuring robust and accurate robot odometry, particularly in perception-degraded environments such as smoke, tunnels, and highly dynamic scenarios.

\section{Approach}

\begin{figure*}
\centering{}
\includegraphics[width=0.92\textwidth]{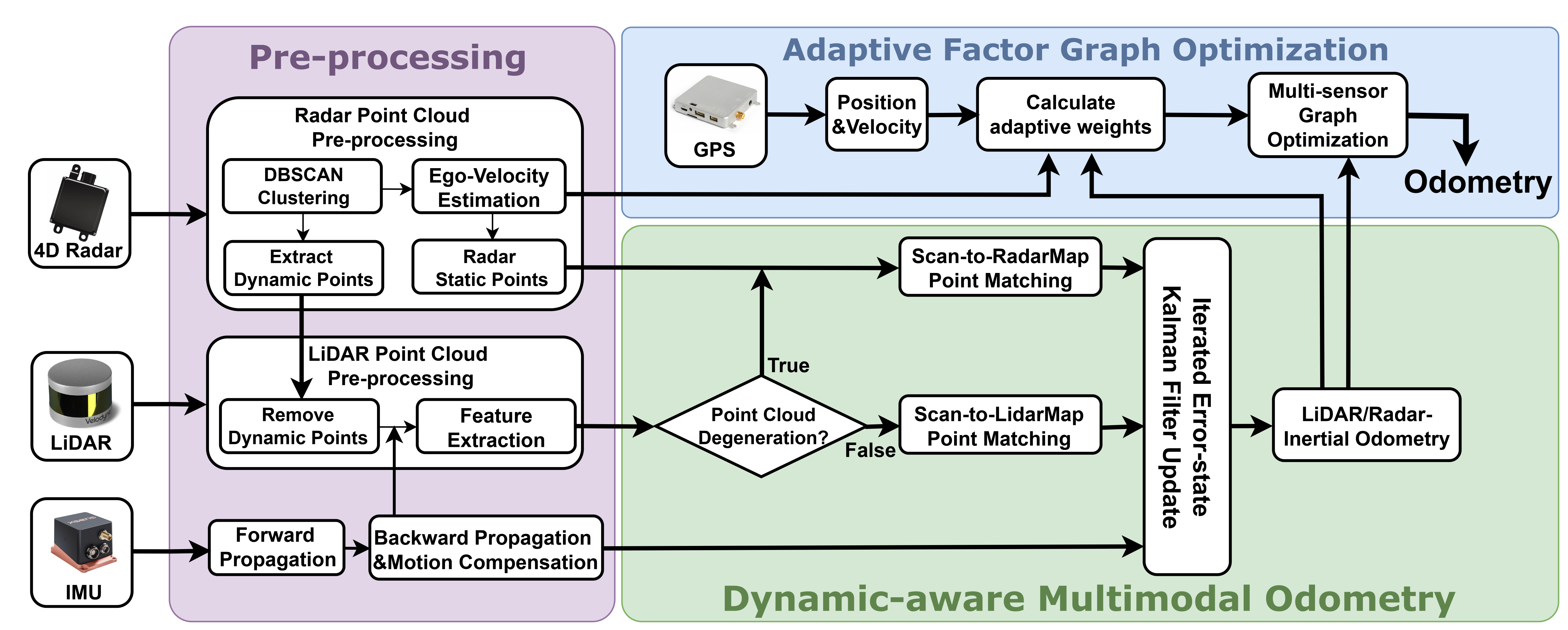}
\caption{Framework overview of AF-RLIO. 
This method can adaptively switch between LiDAR and radar fusion. The multi-modal fusion strategy employs radar point clouds to assist LiDAR in eliminating dynamic obstacles. In environments where LiDAR point clouds are unavailable, the system switches to tightly integrating radar and IMU information to obtain odometry, which is further refined using GPS in backend optimization.}
\label{Framework}
\vspace{-5mm}
\end{figure*}

\subsection{Overview}

The overview of our AF-RLIO system is shown in Fig.~\ref{Framework}, consisting of three main modules: point cloud pre-processing, dynamic-aware multimodal odometry, and adaptive factor graph optimization. The point cloud pre-processing module utilizes 4D millimeter-wave radar, LiDAR, and IMU to extract dynamic and static points from the radar point cloud. The aligned radar point cloud is then used to assist LiDAR in removing dynamic obstacles. The dynamic-aware multimodal odometry module selects the appropriate point cloud data by determining whether the pre-processed LiDAR point cloud has degraded. 
When the robot enters environments where LiDAR becomes ineffective, the system automatically switches to the radar-inertial system to maintain odometry. 
Once the environment returns to normal, the system reverts to the LiDAR-inertial system for SLAM. The factor graph optimization incorporates GPS data, where the GPS factor automatically adjusts its weight based on environmental changes to handle anomalies, ultimately generating a smooth and reliable odometry output.

\subsection{Point Cloud Pre-processing}

\subsubsection{Radar Pre-processing}
The 4D radar provides measurements of 3D positions and Doppler velocities. However, raw 4D radar data are often contaminated by noise and clutter. Initially, the radar's Doppler velocity is used to preliminarily estimate the ego velocity, which aids in distinguishing between dynamic and static point clouds within a single frame. It is important to note that these dynamic points do not necessarily represent dynamic obstacles. They are often generated by moving robots or pedestrians and are characterized by high point cloud density and consistent Doppler information. To further refine the extraction of dynamic point clouds, we employ the density-based spatial clustering algorithm DBSCAN \cite{ester1996density}. This method identifies clusters based on the density of surrounding data points, effectively filtering out noise points and discarding low-density point cloud clusters with inconsistent Doppler information, ensuring the accurate extraction of dynamic robots and pedestrians, as shown in Fig.~\ref{Point_Cloud_Preprocessing}. The radar ego velocity is further refined using a linear least squares approach proposed in \cite{doer2020ekf}, removing false points under the ground and yielding usable static point clouds. 
The point cloud is then segmented into three mutually exclusive subsets: the dynamic point set $(\cdot ) ^D$, the static point set $(\cdot ) ^S$ and the noise point set $(\cdot ) ^N$. Formally, this segmentation is expressed as:
\begin{equation}
{R}= {R}^S \cup  {R}^D \cup  {R}^N,
\end{equation}
where $(\cdot ) ^S \cap (\cdot ) ^D \cap (\cdot ) ^N = \varnothing$.

\subsubsection{LiDAR Pre-processing}

Since LiDAR does not inherently provide velocity information, the presence of dynamic obstacles within the LiDAR point cloud can negatively impact the accuracy of point cloud registration. Consequently, relying solely on LiDAR for the swift and accurate removal of dynamic obstacles is challenging. To address this, our system employs radar to assist in the real-time removal of dynamic points from each LiDAR frame.
In this process, the inputs are the radar dynamic point cloud ${R}^D$ and the LiDAR point cloud $L$. Coordinate alignment is performed to ensure accurate integration of point cloud data from different sensors within a single frame, including spatial calibration and time synchronization. 
High-frequency IMU data are used to align the radar frame with the LiDAR frame through backward propagation. After precise processing, the dynamic points extracted from the radar effectively represent the positions of dynamic obstacles. 
However, achieving precise correspondence between LiDAR and radar points is challenging due to the disparity in point cloud sparsity between the two sensors. 
\begin{figure}

    \centering
    \includegraphics[width=1 \linewidth]{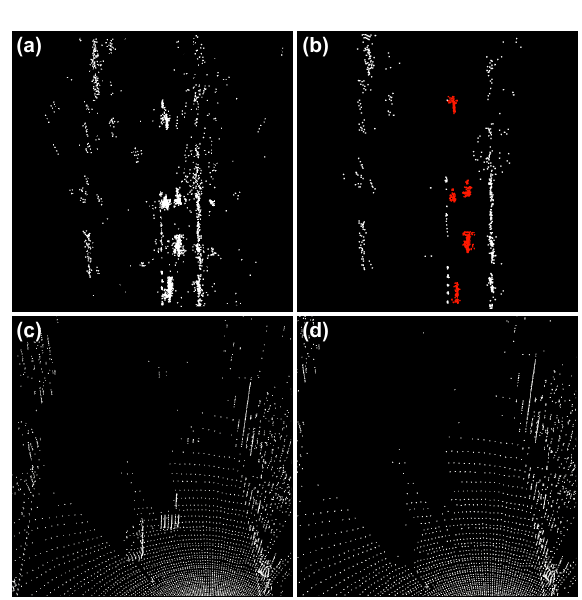}
    \caption{LiDAR and 4D radar point clouds before and after preprocessing. (a) and (c): Raw radar and LiDAR point clouds; (b): Radar point cloud after dynamic-static separation and noise removal; (d): LiDAR point cloud after removing dynamic obstacles.}
    \label{Point_Cloud_Preprocessing}
    \vspace{-4mm}
\end{figure}
To address this, we use a kd-tree structure to organize both point clouds and employ an Euclidean distance-based calculation to determine whether a point in the LiDAR point cloud corresponds to a dynamic point. Specifically, if the Euclidean distance between a LiDAR point ${L}_{i}$ and a dynamic radar point ${R}_{j}^D$ is below a certain threshold, the LiDAR point is classified as dynamic and removed from the LiDAR point cloud. As shown in Fig.~\ref{Point_Cloud_Preprocessing}, the filtered LiDAR point cloud set ${L}_{\text{filtered}}$ that excludes dynamic points can be obtained:

\begin{equation}
{L}_{\text{filtered}}=\left\{{L}_{i}^S\in{L}\left|\min_{j\in\{1,2,\ldots,n\}}d({L}_{i},{R}_{j}^D)\le\epsilon\right\}\right.{,}
\end{equation}
where $d(\cdot )$ denotes the Euclidean distance. $i$ and $j$ represent the LiDAR and radar point cloud sequences, respectively.

\subsection{Dynamic-aware Multimodal Odometry}

We denote the robot state by $\mathbf{x}$, and the state at the $i$-th IMU sampling time is expressed as:

\begin{equation}
\begin{aligned} {\mathbf{x}}&=\begin{bmatrix}{{ }\mathbf {R}{_i}}^T   {{ }\mathbf {p}{_i}}^T  {{ }\mathbf {v}{_i}}^T   {\mathbf{b}{_\omega }}^T {\mathbf {b}{_a}}^T   {{ }\mathbf {g}}^T \end{bmatrix}^T ,
\end{aligned}
\end{equation}
where $\mathbf{R}_{i}$ denotes a rotation matrix on SO(3), $\mathbf{p}_{i}$ and $\mathbf{v}_{i}$ represent the position and velocity, respectively. The terms $\mathbf{b}_{\omega}$ and $\mathbf{b}_{a}$ correspond to the biases from the IMU's gyroscope and accelerometer, respectively. Additionally, $\mathbf{g}$ represents the gravity vector in the world coordinate system, which needs to be estimated. The error state can be expressed in the tangent space near the working point as $ \mathbf{\delta x \in \mathbb{R}^{18}} $. Upon receiving high-frequency IMU data, the current state of the robot is first estimated through forward propagation to obtain the error state $\tilde{\mathbf{x}}_{i+1}$  and covariance $\hat{\mathbf{P}}_{i+1}$.

Compared to scan-to-scan matching, scan-to-map matching offers greater stability and accuracy\cite{xu2022fast}. Therefore, we employ an ikd-tree-based\cite{cai2021ikd} scan-to-map matching approach, which facilitates real-time computation. In typical environments, LiDAR generates point clouds with relatively stable density and high accuracy, making them well-suited for matching with maps. However, in environments like tunnels or smoke-filled areas, the density of LiDAR point clouds decreases, and extracting distinct shape features from the point clouds becomes challenging, making successful matching difficult. Therefore, it is crucial to assess the quality of the point cloud in the current frame before tightly coupling it with the IMU.

\begin{figure}
    \centering
    \includegraphics[width=1 \linewidth]{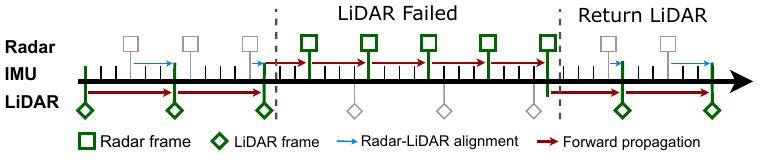}
    \caption{The dynamic perception multi-modal odometry proposed in this paper. As illustrated, IMU measurements serve as the driving force in the architecture, enabling robust odometry through tight coupling with either LiDAR or radar.}
    \label{State_Propagation}
    \vspace{-3mm}
\end{figure}

After removing dynamic points, we extract feature points from the LiDAR data. When the proportion of feature points consistently drops below one percent of the total point cloud, the robot is considered to enter a degraded environment. In such cases, radar point clouds can replace LiDAR point clouds for scan-to-map matching, as shown in Fig.~\ref{State_Propagation}. However, due to differences in point cloud resolution and feature characteristics between radar and LiDAR, direct matching may result in poor robustness and potential system failure from incorrect matches because of the relatively sparse radar point clouds. To address this, before switching from the LiDAR-inertial subsystem to the radar-inertial subsystem, we pre-construct radar submaps and perform radar scan-to-radar submap matching after the switch. Similarly, when the density of LiDAR point clouds and the number of feature points return to normal, we pre-construct LiDAR submaps and transition back from the radar-inertial subsystem to the LiDAR-inertial subsystem. Matching LiDAR scans to LiDAR submaps allows for state updates in the iterative Kalman filter, mitigating jump errors caused by initial matching errors during subsystem transitions.

We employ the IESKF approach to update the observation model iteratively. In each iteration, the current variable $\mathbf{\delta x}$ is calculated and used to derive the next iteration's state vector $\mathbf{x}$ and covariance matrix $\mathbf{P}$, continuing until convergence is achieved. Essentially, each iteration involves solving a least-squares problem with priors:

\begin{equation}
\min_{\widetilde{{\mathbf{x}}}_k^\kappa}\left(\|\mathbf{x}_k\boxminus\widehat{\mathbf{x}}_k\|_{\widehat{\mathbf{P}}_k^{-1}}^2+\sum_{j=1}^m\|r(\widetilde{\mathbf{x}}_k^\kappa)\|_{\mathbf{R}_j^{-1}}^2\right),
\end{equation}
where $k$ denotes the end time of the $j$-th LiDAR scan, $\mathbf{x}_k\boxminus\widehat{\mathbf{x}}_k$ represents the state prediction error, $\widehat{\mathbf{P}}_k$ denotes the propagation covariance, ${r}(\widetilde{\mathbf{x}}_k^\kappa)$ is the observation residual indicating the error between current state estimate and observed data. $\mathbf{R}_j$ is the noise covariance of the $j$-th point.

\subsection{Adaptive Factor Graph Optimization}
GPS can provide accurate position information in open environments but can be interfered with or blocked in scenarios such as tunnels or dense urban areas. 
GPS outliers can also severely distort the fusion-based odometry.
To quickly and effectively detect GPS outliers, we propose a method based on residual $\chi ^2$ testing, which evaluates the reliability of GPS measurements by examining the consistency between the odometry pose data and velocity estimates. By adjusting the GPS weight in the factor graph optimization, the impact of outliers can be mitigated.

We define the pose change $\Delta\mathbf{p}$ and velocity $\mathbf{v}$ between two frames as the consistency criteria for the $\chi ^2$ test. The pose change is derived from the LiDAR/radar-inertial system, while the velocity is estimated using the filtered radar static point cloud ${R}^S$. Thus, we define the residual $\mathbf{e}_k$ as:

\begin{equation}
\mathbf{e}_{k} =\begin{bmatrix}
 \mathbf{e}_{\Delta P} \\\mathbf{e}_{V} 
\end{bmatrix}
=\begin{bmatrix}
 \Delta\mathbf{ p}_{RLIO}^\mathrm{}-\Delta\mathbf{ p}_{GPS}^\mathrm{} \\
 \mathbf{v}_{Radar}^\mathrm{}-\mathbf{v}_{GPS}^\mathrm{} 
\end{bmatrix}.
\end{equation}

If the GPS functions normally, the residual $e_{k}$ should follow a zero-mean Gaussian distribution; otherwise, it will be biased. The decision variable $\mathbf{\lambda }_k$ is defined as the weighted sum of squared residuals, where the weights are provided by the associated covariance matrix $\mathbf{A}$:
\begin{equation}
\mathrm{\lambda }_k=\mathbf{e}_k^T\mathbf{A}^{-1}\mathbf{e}_k .
\end{equation}

Since GPS represents a six-dimensional state, the outlier test function $\mathrm{\lambda }_k$ corresponds to a 6-DoF $\chi ^2$ distribution. After calculating the test statistic  $\mathrm{\lambda }_k$ for the $k$-th frame, we process $\mathrm{\lambda }_k$ as follows:
\begin{equation}
\left.\left\{\begin{array}{ll}\text{Normal,}&\text{if  } \mathrm{\lambda }_k<T_{min}\\
\text{Uncertain,}&\text{if  } T_{min}\le \mathrm{\lambda }_k\le T_{max}\\
\text{Abnormal,}&\text{if } \mathrm{\lambda }_k>T_{max}\end{array}\right..\right.
\end{equation}

When the outlier test function $\mathrm{\lambda }_k$ is less than $T_{min}$, it indicates that the GPS data can be fully trusted. In this case, the factor graph optimization can rely more on GPS data to reduce cumulative error. If the test function exceeds $T_{max}$, it signifies that the GPS data is unreliable, and the optimization will directly remove the GPS factor to eliminate the influence of outliers. When $\mathrm{\lambda}_k$ falls within the intermediate threshold range, the GPS data is considered questionable. In scenarios where GPS data has not been integrated for an extended period, discrepancies may arise between the GPS data and the odometry pose. Directly incorporating the GPS factor under such conditions can cause odometry jumps or even system failure. Therefore, GPS data within the intermediate threshold can serve as a transitional factor in the optimization process, ensuring smooth fusion of odometry while gradually aligning with or diverging from the GPS data as appropriate. When the GPS data falls within the uncertain region, a smoothing function is applied to facilitate this transition:
\begin{equation}
\rho _k=\alpha e_k+(1-\alpha)\rho _{k-1}.
\end{equation}

The smoothing coefficient $\alpha$ (where $0 <\alpha <1$) is used to regulate the transition. When the GPS measurement is considered as a constraint in the pose optimization process, the resulting $\rho _k$ is utilized in the cost function rather than simply minimizing the sum of squared residuals:
\begin{equation}
\mathop{\arg\min}\limits_{\mathbf{x}}\sum  \left \|  r(\mathbf{z}_{RLIO},\mathbf{x})\right \|+\sum \rho_{k}\left \| r(\mathbf{z}_{GPS},\mathbf{x}) \right \|,
\end{equation}
where $r\left ( \cdot  \right ) $ represents the residual of the sensor measurement factor. This approach allows for a gradual integration of GPS data, making the optimization process more robust to potential discrepancies between the GPS and odometry data, thereby maintaining the stability of the overall system.

\section{Experiments}

We evaluated our AF-RLIO system using the MSC Dataset \cite{choi2023msc}, the Snail Dataset \cite{huai2024snail}, and data collected from a real-world smoke environment. We calculated the Root Mean Square Error (RMSE) of Absolute Pose Error (APE) and Relative Pose Error (RPE) as the metrics \cite{grupp2017evo}.

\subsection{MSC Dataset}

For comparison, we selected the widely recognized LiDAR-inertial system FAST-LIO2 \cite{xu2022fast}, the radar-inertial SLAM method RIO, and the LiDAR-radar-inertial fusion method LRIO as baselines. Among them, RIO is derived from our method with LiDAR disabled, while LRIO utilizes both LiDAR and radar simultaneously.

The selected scenarios include U\underline{\hspace{0.5em}}C0, U\underline{\hspace{0.5em}}D0, and U\underline{\hspace{0.5em}}F0, which are characterized by high dynamic obstacles. Additionally, R\underline{\hspace{0.5em}}A2 represents a snowy scene, and U\underline{\hspace{0.5em}}B0 corresponds to a tunnel-crossing scenario. U\underline{\hspace{0.5em}}A0 is a static environment. Our evaluation mainly focuses on the metrics of RPE and APE, using RTK as the ground truth.

\begin{table}[htbp]
\centering
\caption{
\centering {APE (m) / RPE (m) of SLAM systems on MSC Dataset}}
\label{ape_MSC_Dataset}
\resizebox{0.45\textwidth}{!}{
\begin{tabular}{ccccccc}
\toprule
Seq ID & FAST-LIO2\cite{xu2022fast} & RIO& LRIO & \textbf{AF-RLIO} \\
\midrule
U\_{}A0 & \textbf{1.80/0.256} & 9.62/0.357& 1.88/0.268 & \textbf{1.80/0.256} \\
U\_{}B0 & 21.67/* & 9.10/* & 5.05/* & \textbf{2.59/*} \\
U\_{}C0 & 1.30/0.676 & \textbf{1.25}/0.683 & 1.28/0.688 & 1.28/\textbf{0.672} \\
U\_{}D0 & 1.31/1.495 & 2.64/6.701 & 3.70/1.565 & \textbf{1.04/1.490} \\
U\_{}F0 & 1.91/1.202 & 4.42/3.139 & 2.20/\textbf{1.201} & \textbf{1.48}/1.202 \\
R\_{}A2 & 2.81/1.360 &  10.95/4.741 & 11.46/1.378 & \textbf{2.76/1.359} \\
\bottomrule
\end{tabular}
}
\vspace{-3mm}
\end{table}

The pose error results of different algorithms on the MSC dataset are presented in Table \ref{ape_MSC_Dataset}. In these scenarios, the proposed AF-RLIO system enhances point cloud registration accuracy by removing dynamic obstacles, thereby improving localization accuracy in highly dynamic environments. The effectiveness of the AF-RLIO algorithm is particularly evident in tunnel environments, where LiDAR may fail to register due to geometric similarities, causing the failure of LiDAR-inertial odometry. 
In such cases, radar can receive more comprehensive point cloud data to maintain odometry. Compared to using RIO alone, our system leverages LIO in environments outside tunnels with rich geometric features, achieving higher accuracy. Although LRIO can mitigate the degradation effects within tunnels, its accuracy is compromised by the relatively higher errors in radar point clouds, resulting in decreased precision. In predominantly static environments, our algorithm maintains performance comparable to that of FAST-LIO2.

\subsection{Snail Dataset}

We further compared our method with other algorithms on the Snail Dataset. Compared to the MSC Dataset, the Snail Dataset has fewer sequences but has longer sequence distances and includes environments with high dynamics and tunnels, which more effectively highlights the advantages of our algorithm.

\begin{table}[ht]
\centering
\caption{
\centering {APE (m) / RPE (m) of SLAM systems on Snail Dataset}}
\label{ape/rpe_Snail_Dataset}
\resizebox{0.49\textwidth}{!}{
\begin{tabular}{ccccccc}
\toprule
Seq ID & FAST-LIO2\cite{xu2022fast} & RIO & LRIO & \textbf{AF-RLIO}  \\
\midrule
81R&203.3/0.555 &216.7/0.828 &168.0/0.536 &\textbf{24.9/0.387} \\
IAF&41.2/\textbf{0.561} &60.0/1.130 &61.1/0.582 &\textbf{38.7}/0.562\\
IF&15.6/0.566 &23.7/0.838 &17.6/0.567 &\textbf{14.6/0.564}\\
IFEA&16.5/0.374 &34.3/0.949 &22.9/\textbf{0.371 }&\textbf{16.4}/0.373\\
\bottomrule
\end{tabular}
}
\end{table}

The trajectory error results of the Snail dataset using different algorithms are shown in Table \ref{ape/rpe_Snail_Dataset}. The 81R sequence, which includes both tunnel and highway environments, is particularly challenging. Our algorithm demonstrated significant advantages in this sequence, maintaining stable odometry output even under substantial degradation. Although the LRIO algorithm did not experience degradation, continuously incorporating radar data in the optimization process led to negative effects in environments with distinct geometric features. In the IAF sequence, which is characterized by dynamic obstacles, our system effectively leveraged its strengths within the LIO framework. IF and IFEA are campus sequences without high dynamics.

\subsection{Smoke Environment}

We also conducted real-world experiments in smoke environments to validate the feasibility and versatility of the proposed algorithm. The Jackal ground robot was equipped with a Velodyne VLP-16 LiDAR and a 4DAres-R7861 forward radar, both operating at a scanning frequency of 10Hz for acquiring point cloud data. The radar has an elevation angle of $\pm 15^{\circ}$, an azimuth angle of $\pm 75^{\circ}$, an azimuth resolution of $0.25^{\circ}$, an elevation resolution of $0.5^{\circ}$, and a detection range of approximately 200 m. The ground truth trajectory was obtained using RTK-GPS. All computations were performed on a laptop with Ubuntu 20.04 and ROS Noetic. All tests were conducted on an 24-core Intel i7-14650HX CPU. The robot and sensor configuration are shown in Fig. \ref{Real_Experiment}.

\begin{table}[htbp]
\centering
\caption{\centering {Comparison of APE (m) / RPE (m) in the smoke environment}}
\label{ape/rpe_Smoke_Environment}
   \resizebox{0.45\textwidth}{!}{
\begin{tabular}{@{}cccccc@{}}
\toprule
\textbf{ }  & FAST-LIO2\cite{xu2022fast} & RIO & LRIO & \textbf{AF-RLIO} \\ \midrule
SMOKE & 12.8/0.141 & 0.67/0.116 & 0.63/0.115 &\textbf{0.55/0.114} \\ \bottomrule
\end{tabular}
}
\vspace{-4mm}
\end{table}

\begin{figure}[ht]
    \centering
    \includegraphics[width=0.99\linewidth]{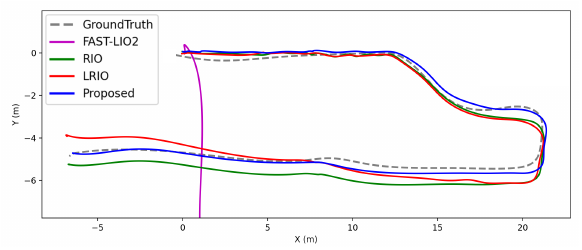}
    \caption{Comparative trajectories in real-world smoke environments.}
    \label{Smoke_Environment}
\end{figure}

We collected a dataset of the robot navigating in an outdoor area where smoke generators were deployed to create dense smoke, significantly reducing the effective range of the LiDAR and obscuring large portions of its field of view. Even in areas not fully obstructed by the smoke, the LiDAR struggled to capture accurate point cloud data. As the smoke dissipated, the LiDAR gradually regained reliable point cloud data. 
The trajectory comparison between the proposed method and other SLAM algorithms in smoke environments is illustrated in Fig.~\ref{Smoke_Environment}, and pose errors are presented in Table~\ref{ape/rpe_Smoke_Environment}. It can be seen that FAST-LIO2 immediately failed in the presence of smoke due to the high noise and limited range of the LiDAR under such conditions. While RIO and LRIO maintained the pose relatively well, they were less accurate than AF-RLIO because the errors in the radar data interfered with the final pose estimation in smoke-free environments.

\subsection{GPS-Challenge Environment Tests}

The URBAN\_B0 sequence in the MSC dataset involves passing through a tunnel where GPS signals are occluded, making accurate localization difficult. As the robot enters the tunnel, data uncertainty increases with depth, but upon exiting into a more open environment, GPS data gradually becomes more accurate.

\begin{table}[htbp]
\centering
\caption{\centering {Comparison of APE (m) in the GPS-challenge environment}}
\label{ape_GPS_Challenge_Environment}
   \resizebox{0.45\textwidth}{!}{
\begin{tabular}{@{}cccccc@{}}
\toprule
\textbf{ }  & Constant-GPS & Threshold-GPS & \textbf{Adaptive-GPS} \\ \midrule
URBAN\_{}B0 & failed & 22.82 &\textbf{2.26} \\ \bottomrule
\end{tabular}
}
\vspace{-1mm}
\end{table}

\begin{figure}[ht]
    \centering
    \includegraphics[width=0.99\linewidth]{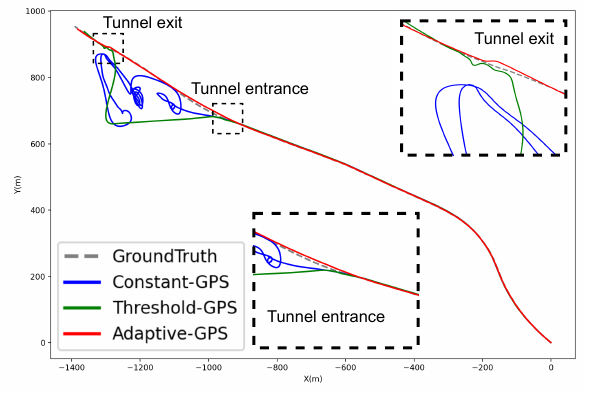}
    \caption{Trajectory comparison of no GPS outlier detection (Constant-GPS), threshold-based outlier detection method (Threshold-GPS), and the proposed outlier detection method (Adaptive-GPS) in the tunnel scenario.}
    \label{GPS_Challenge_Environment}
    \vspace{-4mm}
\end{figure}

We compared three algorithms: Proposed-Constant without outlier detection, Proposed-Threshold based on threshold judgment, and an adaptive algorithm. The resulting trajectories are shown in Fig.\ref{GPS_Challenge_Environment}, and the APE results are presented in Table \ref{ape_GPS_Challenge_Environment}. In the algorithm without outlier detection, the system fails due to the inclusion of erroneous GPS data. The threshold-based GPS judgment fails as it cannot optimally determine when to accept or reject GPS data, leading to divergence in optimization. The adaptive method, however, quickly and accurately assesses the reliability of GPS data, reducing its weight in the optimization as uncertainty increases, ensuring that odometry is not affected by erroneous data. Additionally, as the robot exits the tunnel, the adaptive method smoothly increases the GPS weight, mitigating data jumps during optimization. The results demonstrate that our method effectively detects and removes GPS outliers, suppresses error accumulation, and avoids the impact of GPS anomalies.

\section{Conclusions}

This paper presents a multimodal fusion-based odometry, AF-RLIO, that integrates IMU and GPS with adaptively-switching LiDAR and radar sensors. It ensures robust odometry for robots in challenging environments such as tunnels, smoke-filled areas, and indoor-outdoor transitions. By leveraging the complementary advantages of radar and LiDAR point clouds, AF-RLIO employs an IESKF-based method to tightly couple pre-selected points with IMU data, thereby achieving accurate odometry. In the back end, adaptively weighted GPS data is incorporated into the joint optimization with the odometry, effectively reducing accumulated errors during long-term mapping and facilitating a smooth transition in the presence of intermittent GPS errors. The algorithm has been tested on multiple datasets and in real-world scenarios, demonstrating strong robustness in environments with similar geometric features and under smoke conditions.


{
\bibliographystyle{ieeetr}
\bibliography{afrlio}

\begin{thebibliography}{10}

\bibitem{ebadi2022present}
K.~Ebadi, L.~Bernreiter, H.~Biggie, G.~Catt, Y.~Chang, A.~Chatterjee, C.~E. Denniston, S.-P. Desch{\^e}nes, K.~Harlow, S.~Khattak, {\em et~al.}, ``Present and future of slam in extreme underground environments,'' {\em arXiv preprint arXiv:2208.01787}, 2022.

\bibitem{lee2024lidar}
D.~Lee, M.~Jung, W.~Yang, and A.~Kim, ``Lidar odometry survey: recent advancements and remaining challenges,'' {\em Intelligent Service Robotics}, vol.~17, no.~2, pp.~95--118, 2024.

\bibitem{bijelic2018benchmark}
M.~Bijelic, T.~Gruber, and W.~Ritter, ``A benchmark for lidar sensors in fog: Is detection breaking down?,'' in {\em 2018 IEEE intelligent vehicles symposium (IV)}, pp.~760--767, IEEE, 2018.

\bibitem{liu2021optimization}
J.~Liu, W.~Gao, and Z.~Hu, ``Optimization-based visual-inertial slam tightly coupled with raw gnss measurements,'' in {\em 2021 IEEE International Conference on Robotics and Automation (ICRA)}, pp.~11612--11618, IEEE, 2021.

\bibitem{nubert2022graph}
J.~Nubert, S.~Khattak, and M.~Hutter, ``Graph-based multi-sensor fusion for consistent localization of autonomous construction robots,'' in {\em 2022 International Conference on Robotics and Automation (ICRA)}, pp.~10048--10054, IEEE, 2022.

\bibitem{wang2024end}
R.~Wang, Y.~Jing, C.~Gu, S.~He, and J.~Chen, ``End-to-end multi-target flexible job shop scheduling with deep reinforcement learning,'' {\em IEEE Internet of Things Journal}, 2024.

\bibitem{harlow2023new}
K.~Harlow, H.~Jang, T.~D. Barfoot, A.~Kim, and C.~Heckman, ``A new wave in robotics: Survey on recent mmwave radar applications in robotics,'' {\em arXiv preprint arXiv:2305.01135}, 2023.

\bibitem{qin2020lins}
C.~Qin, H.~Ye, C.~E. Pranata, J.~Han, S.~Zhang, and M.~Liu, ``Lins: A lidar-inertial state estimator for robust and efficient navigation,'' in {\em 2020 IEEE international conference on robotics and automation (ICRA)}, pp.~8899--8906, IEEE, 2020.

\bibitem{xu2021fast}
W.~Xu and F.~Zhang, ``Fast-lio: A fast, robust lidar-inertial odometry package by tightly-coupled iterated kalman filter,'' {\em IEEE Robotics and Automation Letters}, vol.~6, no.~2, pp.~3317--3324, 2021.

\bibitem{xu2022fast}
W.~Xu, Y.~Cai, D.~He, J.~Lin, and F.~Zhang, ``Fast-lio2: Fast direct lidar-inertial odometry,'' {\em IEEE Transactions on Robotics}, vol.~38, no.~4, pp.~2053--2073, 2022.

\bibitem{wu2024lio}
Y.~Wu, T.~Guadagnino, L.~Wiesmann, L.~Klingbeil, C.~Stachniss, and H.~Kuhlmann, ``Lio-ekf: High frequency lidar-inertial odometry using extended kalman filters,'' in {\em 2024 IEEE International Conference on Robotics and Automation (ICRA)}, pp.~13741--13747, IEEE, 2024.

\bibitem{bai2022faster}
C.~Bai, T.~Xiao, Y.~Chen, H.~Wang, F.~Zhang, and X.~Gao, ``Faster-lio: Lightweight tightly coupled lidar-inertial odometry using parallel sparse incremental voxels,'' {\em IEEE Robotics and Automation Letters}, vol.~7, no.~2, pp.~4861--4868, 2022.

\bibitem{ye2019tightly}
H.~Ye, Y.~Chen, and M.~Liu, ``Tightly coupled 3d lidar inertial odometry and mapping,'' in {\em 2019 International Conference on Robotics and Automation (ICRA)}, pp.~3144--3150, IEEE, 2019.

\bibitem{shan2020lio}
T.~Shan, B.~Englot, D.~Meyers, W.~Wang, C.~Ratti, and D.~Rus, ``Lio-sam: Tightly-coupled lidar inertial odometry via smoothing and mapping,'' in {\em 2020 IEEE/RSJ international conference on intelligent robots and systems (IROS)}, pp.~5135--5142, IEEE, 2020.

\bibitem{li2021towards}
K.~Li, M.~Li, and U.~D. Hanebeck, ``Towards high-performance solid-state-lidar-inertial odometry and mapping,'' {\em IEEE Robotics and Automation Letters}, vol.~6, no.~3, pp.~5167--5174, 2021.

\bibitem{zhao2021super}
S.~Zhao, H.~Zhang, P.~Wang, L.~Nogueira, and S.~Scherer, ``Super odometry: Imu-centric lidar-visual-inertial estimator for challenging environments,'' in {\em 2021 IEEE/RSJ International Conference on Intelligent Robots and Systems (IROS)}, pp.~8729--8736, IEEE, 2021.

\bibitem{ebadi2023present}
K.~Ebadi, L.~Bernreiter, H.~Biggie, G.~Catt, Y.~Chang, A.~Chatterjee, C.~E. Denniston, S.-P. Desch{\^e}nes, K.~Harlow, S.~Khattak, {\em et~al.}, ``Present and future of slam in extreme environments: The darpa subt challenge,'' {\em IEEE Transactions on Robotics}, 2023.

\bibitem{li20234d}
X.~Li, H.~Zhang, and W.~Chen, ``4d radar-based pose graph slam with ego-velocity pre-integration factor,'' {\em IEEE Robotics and Automation Letters}, 2023.

\bibitem{lim2023orora}
H.~Lim, K.~Han, G.~Shin, G.~Kim, S.~Hong, and H.~Myung, ``Orora: Outlier-robust radar odometry,'' in {\em 2023 IEEE International Conference on Robotics and Automation (ICRA)}, pp.~2046--2053, IEEE, 2023.

\bibitem{burnett2021we}
K.~Burnett, A.~P. Schoellig, and T.~D. Barfoot, ``Do we need to compensate for motion distortion and doppler effects in spinning radar navigation?,'' {\em IEEE Robotics and Automation Letters}, vol.~6, no.~2, pp.~771--778, 2021.

\bibitem{huai2024snail}
J.~Huai, B.~Wang, Y.~Zhuang, Y.~Chen, Q.~Li, Y.~Han, and C.~Toth, ``Snail-radar: A large-scale diverse dataset for the evaluation of 4d-radar-based slam systems,'' {\em arXiv preprint arXiv:2407.11705}, 2024.

\bibitem{choi2023msc}
M.~Choi, S.~Yang, S.~Han, Y.~Lee, M.~Lee, K.~H. Choi, and K.-S. Kim, ``Msc-rad4r: Ros-based automotive dataset with 4d radar,'' {\em IEEE Robotics and Automation Letters}, 2023.

\bibitem{cen2018precise}
S.~H. Cen and P.~Newman, ``Precise ego-motion estimation with millimeter-wave radar under diverse and challenging conditions,'' in {\em 2018 IEEE International Conference on Robotics and Automation (ICRA)}, pp.~6045--6052, IEEE, 2018.

\bibitem{cen2019radar}
S.~H. Cen and P.~Newman, ``Radar-only ego-motion estimation in difficult settings via graph matching,'' in {\em 2019 International Conference on Robotics and Automation (ICRA)}, pp.~298--304, IEEE, 2019.

\bibitem{lu2020see}
C.~X. Lu, S.~Rosa, P.~Zhao, B.~Wang, C.~Chen, J.~A. Stankovic, N.~Trigoni, and A.~Markham, ``See through smoke: robust indoor mapping with low-cost mmwave radar,'' in {\em Proceedings of the 18th International Conference on Mobile Systems, Applications, and Services}, pp.~14--27, 2020.

\bibitem{guan2020through}
J.~Guan, S.~Madani, S.~Jog, S.~Gupta, and H.~Hassanieh, ``Through fog high-resolution imaging using millimeter wave radar,'' in {\em Proceedings of the IEEE/CVF Conference on Computer Vision and Pattern Recognition}, pp.~11464--11473, 2020.

\bibitem{kubelka2024we}
V.~Kubelka, E.~Fritz, and M.~Magnusson, ``Do we need scan-matching in radar odometry?,'' in {\em 2024 IEEE International Conference on Robotics and Automation (ICRA)}, pp.~13710--13716, IEEE, 2024.

\bibitem{michalczyk2022tightly}
J.~Michalczyk, R.~Jung, and S.~Weiss, ``Tightly-coupled ekf-based radar-inertial odometry,'' in {\em 2022 IEEE/RSJ International Conference on Intelligent Robots and Systems (IROS)}, pp.~12336--12343, IEEE, 2022.

\bibitem{doer2020ekf}
C.~Doer and G.~F. Trommer, ``An ekf based approach to radar inertial odometry,'' in {\em 2020 IEEE International Conference on Multisensor Fusion and Integration for Intelligent Systems (MFI)}, pp.~152--159, IEEE, 2020.

\bibitem{michalczyk2023multi}
J.~Michalczyk, R.~Jung, C.~Brommer, and S.~Weiss, ``Multi-state tightly-coupled ekf-based radar-inertial odometry with persistent landmarks,'' in {\em 2023 IEEE International Conference on Robotics and Automation (ICRA)}, pp.~4011--4017, IEEE, 2023.

\bibitem{kramer2020radar}
A.~Kramer, C.~Stahoviak, A.~Santamaria-Navarro, A.-A. Agha-Mohammadi, and C.~Heckman, ``Radar-inertial ego-velocity estimation for visually degraded environments,'' in {\em 2020 IEEE International Conference on Robotics and Automation (ICRA)}, pp.~5739--5746, IEEE, 2020.

\bibitem{kramer2021radar}
A.~Kramer and C.~Heckman, ``Radar-inertial state estimation and obstacle detection for micro-aerial vehicles in dense fog,'' in {\em Experimental Robotics: The 17th International Symposium}, pp.~3--16, Springer, 2021.

\bibitem{fritsche2018fusing}
P.~Fritsche, S.~Kueppers, G.~Briese, and B.~Wagner, ``Fusing lidar and radar data to perform slam in harsh environments,'' in {\em Informatics in Control, Automation and Robotics: 13th International Conference, ICINCO 2016 Lisbon, Portugal, 29-31 July, 2016}, pp.~175--189, Springer, 2018.

\bibitem{yin2021rall}
H.~Yin, R.~Chen, Y.~Wang, and R.~Xiong, ``Rall: end-to-end radar localization on lidar map using differentiable measurement model,'' {\em IEEE Transactions on Intelligent Transportation Systems}, vol.~23, no.~7, pp.~6737--6750, 2021.

\bibitem{mielle2019comparative}
M.~Mielle, M.~Magnusson, and A.~J. Lilienthal, ``A comparative analysis of radar and lidar sensing for localization and mapping,'' in {\em 2019 European Conference on Mobile Robots (ECMR)}, pp.~1--6, IEEE, 2019.

\bibitem{nissov2024degradation}
M.~Nissov, N.~Khedekar, and K.~Alexis, ``Degradation resilient lidar-radar-inertial odometry,'' {\em arXiv preprint arXiv:2403.05332}, 2024.

\bibitem{noh2024adaptivelidarradarfusionoutdoor}
C.~Noh and A.~Kim, ``Adaptive lidar-radar fusion for outdoor odometry across dense smoke conditions,'' 2024.

\bibitem{ester1996density}
M.~Ester, H.-P. Kriegel, J.~Sander, X.~Xu, {\em et~al.}, ``A density-based algorithm for discovering clusters in large spatial databases with noise,'' in {\em kdd}, vol.~96, pp.~226--231, 1996.

\bibitem{cai2021ikd}
Y.~Cai, W.~Xu, and F.~Zhang, ``ikd-tree: An incremental kd tree for robotic applications,'' {\em arXiv preprint arXiv:2102.10808}, 2021.

\bibitem{grupp2017evo}
M.~Grupp, ``evo: Python package for the evaluation of odometry and slam..'' \url{https://github.com/MichaelGrupp/evo}, 2017.

\end{thebibliography}
}


\end{document}